\documentclass[runningheads]{llncs}
\usepackage[T1]{fontenc}
\usepackage{graphicx}
\usepackage{amsmath}

\usepackage{stmaryrd}

\usepackage{booktabs}
\usepackage{hyperref}

\begin{document}
\title{Constrained Entity Selection under Partial Knowledge for LLM-Based Knowledge Graph QA\thanks{This preprint has not undergone peer review or any post-submission improvements or corrections. The Version of Record of this contribution is published in KI 2026: Advances in Artificial Intelligence, KI 2026, LNCS, vol 16830, and is available online at \url{https://doi.org/10.1007/978-3-032-32335-4_21}}}
\titlerunning{Constrained Entity Selection under Partial Knowledge}
%
\author{Emanuel Kitzelmann\inst{1}\orcidID{0000-0001-5614-3463}}
\authorrunning{E. Kitzelmann}
%
\institute{Brandenburg University of Applied Sciences, Brandenburg a.d.H., Germany\\
\email{emanuel.kitzelmann@th-brandenburg.de}}
\maketitle              
\begin{abstract}
Large language models are increasingly used for knowledge graph question answering (KGQA), but can fail to correctly ground answers in the underlying graph.
Current approaches to LLM-based KGQA either rely on full semantic parsing into executable queries such as SPARQL, which is brittle in practice due to complex schemas or incompleteness of real-world KGs, or on LLM-reasoning and answer generation over KGs, which can be more robust but lacks formal guarantees.

In this work, we study a complementary setting in which \emph{candidate} answers are generated by an LLM-based system and subsequently verified using lightweight symbolic constraints derived from the question. We introduce \emph{Constrained Entity Selection under Partial Knowledge (CES-PK)}, a problem formulation that focuses on eliminating invalid answers and providing symbolic support for valid ones without requiring construction of executable logical forms. To account for incomplete KGs, we employ a three-valued constraint semantics (\emph{satisfied, violated, unknown}) that avoids incorrect rejections under open-world assumptions.

To demonstrate the effects of our method, we instantiate this framework over the Hetionet biomedical knowledge graph and evaluate the impact of type, relation, and exclusion constraints. Experiments show that precision improves by filtering invalid candidates, while recall is preserved due to retaining candidates whose constraints are not explicitly violated. Satisfied constraints provide additional positive symbolic evidence to rank remaining candidates.

\end{abstract}

\section{Introduction}

Large language models (LLMs) are increasingly used for knowledge graph question answering (KGQA) \cite{MaEtAl2025}, enabling natural language access to structured knowledge. However, LLMs can fail to correctly ground answers in the underlying graph and may violate basic structural or semantic constraints.

Current LLM approaches to KGQA typically follow one of two paradigms. In semantic parsing methods \cite{DAbramoEtAl2025,BhandiwadEtAl2026}, questions are translated into executable logical forms such as SPARQL queries. While this enables precise and verifiable answers, generating correct queries remains challenging in practice, especially for complex schemas and incomplete KGs \cite{ElKhatib2025}. In contrast, retrieval-augmented or agentic approaches such as \cite{SunEtAl2023} use LLMs to retrieve and reason over KGs. These methods are more flexible and robust to missing information but lack formal guarantees and may produce hallucinated or non-grounded answers.

In this work, we propose a complementary perspective that avoids both full semantic parsing and unconstrained LLM reasoning. Instead of deriving complete logical forms, we extract lightweight necessary constraints from the question and use them to verify candidate answers produced by an upstream system. This yields a verification layer that filters invalid candidates, retains candidates that cannot be disproven under incomplete knowledge, and provides symbolic support for valid candidates.

We formalize this setting as \emph{Constrained Entity Selection under Partial Knowledge} (CES-PK). To account for the open-world nature of real-world knowledge graphs, we employ a three-valued constraint semantics that distinguishes between satisfied, violated, and unknown constraints. This allows us to reject candidates only in the presence of explicit violations, thereby avoiding false negatives caused by missing information.

We instantiate this framework for entity-centric queries over the Hetionet\footnote{\url{https://het.io/}} biomedical knowledge graph \cite{HimmelsteinEtAl2017} and evaluate its effect on simulated candidate answer sets. Using controlled candidate sets that simulate LLM-generated answers, we show that constraint-based verification improves precision substantially while preserving recall and yielding support for KG-grounded answers. Furthermore, we analyze the complementary roles of constraint types: exclusion constraints contribute to filtering, whereas positive relation constraints contribute to verification through support.

\paragraph{Contributions.} This paper makes the following contributions:
\begin{itemize}
    \item We introduce Constrained Entity Selection under Partial Knowledge (CES-PK), a problem formulation for post-hoc verification of KGQA answers without requiring full semantic parsing.
    \item We analyze the complementary roles of constraint types for filtering and verification with a three-valued constraint semantics for incomplete KGs.
    \item We provide an empirical study demonstrating that lightweight constraint-based verification improves precision while preserving recall and additionally provides support scores for retained answers.
\end{itemize}

\section{Problem and Method}

\subsection{Problem Setting}

We consider knowledge graph question answering (KGQA) in a setting where candidate answers are produced by an upstream component (e.g., an agentic LLM pipeline) and may contain incorrect or unsupported entities.

Let $G = (V, E)$ be a knowledge graph \cite{HoganEtAl2021} with entities $V$ and typed relations $E \subseteq V \times R \times V$. Given a question $q$, we assume a candidate set $C(q) \subseteq V$ and an (unknown) set of correct answers $A(q) \subseteq V$.
Rather than deriving $A(q)$ directly, we aim to identify candidates consistent with necessary conditions implied by the question.

\paragraph{Constrained Entity Selection under Partial Knowledge (CES-PK).}
Given $C(q)$ and a set of constraints $\mathcal{C}(q)$ derived from $q$, compute a subset $C^*(q) \subseteq C(q)$ such that no retained candidate violates any constraint under incomplete knowledge and associate remaining answers in $C^*(q)$ with support scores reflecting the degree of constraint satisfaction.

\subsection{Assumptions on the Knowledge Graph}

We assume that the KG is factually correct but may be incomplete, i.e., all represented triples are true, while some true facts may be missing (open-world assumption). Furthermore, all candidate entities are assumed to be present in the graph.

\subsection{Constraints}

We consider lightweight necessary constraints extracted from the question with respect to a known schema (this set of constraint types can be extended, e.g., to capture joins or cardinality constraints):

\begin{itemize}
    \item Type constraints: $\text{type}(x, T)$
    \item Relation constraints: $R(x, y)$
    \item Exclusion constraints: $\neg R(x, y)$
\end{itemize}

\subsection{Three-Valued Constraint Semantics}

Due to incompleteness, constraints $c(x)$ for candidate answers $x$ over graph $G$ are evaluated under a three-valued semantics:
\(
\llbracket c(x) \rrbracket_G \in \{\mathsf{sat}, \mathsf{viol}, \mathsf{unk}\}.
\)

\paragraph{Relation constraints}
\[
\llbracket R(x, y) \rrbracket_G =
\begin{cases}
\mathsf{sat} & \text{if } (x, R, y) \in E \\
\mathsf{unk} & \text{otherwise}
\end{cases}
\]

\paragraph{Exclusion constraints}
\[
\llbracket \neg R(x, y) \rrbracket_G =
\begin{cases}
\mathsf{viol} & \text{if } (x, R, y) \in E \\
\mathsf{unk} & \text{otherwise}
\end{cases}
\]

\paragraph{Type constraints}
\[
\llbracket \text{type}(x, T) \rrbracket_G =
\begin{cases}
\mathsf{sat} & \text{if } \text{type}(x) = T \\
\mathsf{viol} & \text{otherwise}
\end{cases}
\]

\subsection{Constraint-Based Filtering and Support-Based Verification}

We retain only candidates that do not violate any constraint:
\[
C^*(q) = \{ x \in C(q) \mid \forall c \in \mathcal{C}(q): \llbracket c(x) \rrbracket_G \neq \mathsf{viol} \}.
\]

This strategy relies solely on explicit violations and is therefore robust under incomplete knowledge.

Further, let
\[
\mathcal{C}^{+}(q) = \{ c \in \mathcal{C}(q) \mid \mathsf{sat} \in \mathrm{range}(c) \}
\]
denote the subset of constraints that can provide positive evidence by being satisfied.
The support of a candidate $x$ is defined as the fraction of constraints in $\mathcal{C}^{+}(q)$ that evaluate to $\mathsf{sat}$:
\[
\mathrm{support}(x, q) =
\frac{
|\{ c \in \mathcal{C}^{+}(q) \mid \llbracket c(x) \rrbracket_G = \mathsf{sat} \}|
}{
|\mathcal{C}^{+}(q)|
}.
\]

\section{Experiments}

\subsection{Experimental Setup}

We evaluate the proposed constraint-based verification approach on entity-centric KGQA tasks using the Hetionet biomedical knowledge graph \cite{HimmelsteinEtAl2017}.\footnote{Code and experimental artifacts: \url{https://github.com/ekitzelmann/ces-pk}} This evaluation is designed as a controlled study to isolate the effect of constraint-based verification independently of upstream candidate generation. Therefore, we simulate answer generation by sampling queries and gold answers from the Hetionet KG and then generate candidate answers from gold answers by systematically adding incorrect answers (false positives). Since our method only removes candidates with explicit constraint violations, recall is unaffected and we do not simulate false negatives.

We consider two query types that fit our constraints (these can be extended by more complex forms including more complex constraint types):
\begin{itemize}
    \item Q1: Which compounds treat disease $D$?
    \item Q2: Which compounds treat disease $D$ but do not cause side effect $S$?
\end{itemize}

For each query type, we derive queries by sampling diseases and side effects from the KG and construct gold answer sets from the graph.
We construct candidate sets that simulate LLM-generated answers by combining gold answers with sampled distractors, including type-incorrect entities and candidates that violate the exclusion constraint (for Q2).

Constraints include type (C1), relation (C2), and exclusion (C3) constraints. Evaluation is based on precision before and after filtering and support scores after filtering. Since all gold answers are included in the candidate sets, recall remains constant at 1.0. In the present experiments, $\mathcal{C}^{+}(q)$ consists of the type and positive relation constraints, such that $|\mathcal{C}^{+}(q)|=2$ for both Q1 and Q2.

\subsection{Results}

\begin{table}[t]
\centering
\caption{Performance of constraint-based filtering and support-based verification.}
\label{tab:results}
\begin{tabular}{lcccccc}
\toprule
Query & Prec. (before) & Prec. (after) & $\Delta$ & Recall & Supp. (gold) & Supp. (non-gold) \\
\midrule
Q1 & 0.66 & 0.70 & +0.04 & 1.00 & 1.00 & 0.42 \\
Q2 & 0.17 & 0.62 & +0.45 & 1.00 & 1.00 & 0.74 \\
\bottomrule
\end{tabular}
\end{table}

\begin{table}[t]
\centering
\caption{Effect of simulated KG incompleteness (10\% edge removal).}
\label{tab:resultsincomp}
\begin{tabular}{lccccc}
\toprule
Query & Prec. (after) & $\Delta$ & Recall & Supp. (gold) & Supp. (non-gold) \\
\midrule
Q1 & 0.70 & +0.04 & 1.00 & 0.95 & 0.42 \\
Q2 & 0.53 & +0.37 & 1.00 & 0.95 & 0.69 \\
\bottomrule
\end{tabular}
\end{table}

Table~\ref{tab:results} summarizes filtering performance and support-based verification evidence on the Hetionet KG.
For Q1, constraint-based filtering yields a modest precision improvement (+0.04) due to removed type-inconsistent candidates. For Q2, we achieve a substantial precision increase from 0.17 to 0.62 (+0.45) due to exclusion constraints that eliminate candidates violating the negated condition. Overall, precision improves from 0.41 to 0.66 while recall remains unaffected.

The results show distinct roles of constraint types. Type constraints (C1) provide consistent filtering 
signals, while exclusion constraints (C3) are the primary driver of filtering in Q2. Relation constraints (C2) do not lead to violations under OWA but contribute to verification by providing support if satisfied.

Support scores provide additional discrimination between candidates. 
Gold answers achieve a mean support of 1.0 compared to 0.42 (Q1) and 0.74 (Q2) for non-gold candidates.
This indicates that positive constraints contribute useful verification signals even when filtering is limited.

To further analyze the impact of KG incompleteness, in a second experiment we randomly removed 10\% of the relevant \textit{treats} and \textit{causes} edges after constructing the gold answer sets. Results are shown in Table~\ref{tab:resultsincomp}. 
Precision and support scores decrease, as missing edges reduce the number of detectable violations and verifications, while the method continues to preserve recall.

\subsection{Discussion}

The results show that lightweight constraints improve precision without requiring full semantic parsing. Negative constraints provide filtering signals when explicit violations are present, while positive constraints contribute evidence for verification. The three-valued semantics preserves recall under incomplete knowledge.

\paragraph{Limitations.}
The evaluation is based on controlled candidate sets rather than capturing end-to-end KGQA performance. The KG is assumed to be factually correct but incomplete. In the presence of incorrect triples or missing entities, constraint evaluation may lead to false rejections and support. 

\section{Related Work}

KGQA with LLMs \cite{BaekEtAl2023,SunEtAl2023,WagnerEtAl2025} is approached in different ways:
In semantic parsing approaches \cite{BerantEtAl2013}, an LLM translates natural language questions into executable logical forms \cite{FengHe2025,DAbramoEtAl2025}.
Such methods enable precise and verifiable answers, but remain brittle in practice as generating correct queries over complex schemas is challenging. They typically involve fine-tuning, require complete KGs and leave LLM knowledge unexploited that could help to reason more robustly over incomplete KGs.

Another line of work uses LLMs for reasoning over knowledge graphs and generating answers. In \cite{BaekEtAl2023}, relevant triples are retrieved from the KG by entity linking which are then verbalized and injected as additional context into the prompt. Think-on-Graph \cite{SunEtAl2023} allows LLMs to iteratively explore graph neighborhoods and construct reasoning paths. These methods typically work without additional training and are more robust but can produce hallucinated or non-grounded answers. Moreover, incomplete KGs remain challenging. To address the issue of unreliable LLM-based reasoning, \cite{LiEtAl2025} introduce constraints constructed from the KG that guide the LLM decoding to generate well-formed reasoning chains. In contrast, the method proposed in this paper uses constraints to post-hoc verify answer candidates. In \cite{ElKhatib2025}, integration of link prediction into LLM-based reasoning over KGs is suggested to address incomplete KGs. 

Our work is also related to research on generation followed by verification \cite{ValmeekamEtAl2023}, where it is shown that LLMs are effective at generating solution candidates but unreliable at verifying them.

\section{Conclusions}

We introduced Constrained Entity Selection under Partial Knowledge (CES-PK), a framework for post-hoc verification of candidate answers in knowledge graph question answering. Instead of relying on full semantic parsing or unconstrained neural generation, our approach uses lightweight necessary constraints together with a three-valued semantics to filter invalid candidates and provide symbolic support for valid ones under incomplete knowledge.

Our experimental results demonstrate that constraint-based verification can substantially improve precision while preserving recall in a controlled setting. In particular, we show that exclusion constraints provide strong negative signals for filtering, whereas positive relation constraints contribute to verification through support.

The proposed approach assumes that the underlying knowledge graph is factually correct but incomplete and that all candidate entities are present in the graph. Furthermore, our evaluation is based on controlled candidate sets and does not yet capture end-to-end KGQA performance under noisy knowledge graphs.

Future work includes integrating the proposed verification layer into full KGQA pipelines, studying robustness under incomplete and noisy graphs, and extending constraint extraction to more complex query types and schemas.


\end{document}